# Sentiment analysis of preservice teachers' reflections using a large language model

Yunsoo Park
Department of Electrical and
Computer Engineering
Iowa State University
Ames(IA), U.S.A.
yunsoopk@iastate.edu

Younkyung Hong
Department of Elementary
Education
Ball State University
Muncie(IN), U.S.A.
yhong@bsu.edu

*Abstract*—In this study, the emotion and tone of preservice teachers' reflections were analyzed using sentiment analysis with LLMs: GPT-4, Gemini, and BERT. We compared the results to understand how each tool categorizes and describes individual reflections and multiple reflections as a whole. This study aims to explore ways to bridge the gaps between qualitative, quantitative, and computational analyses of reflective practices in teacher education. This study finds that to effectively integrate LLM analysis into teacher education, developing an analysis method and result format that are both comprehensive and relevant for preservice teachers and teacher educators is crucial.

## I. Introduction

In teacher education, field experiences that preservice teachers have in classrooms provide them with opportunities to apply their learned content and pedagogical knowledge and to develop professional dispositions by teaching and working in real-world settings [1], [2], [3]. Responding to the call for increasing field experiences, teacher preparation programs around the world have extended the time that preservice teachers spend in classrooms [4]. However, as Zeichner [5] points out, the increased quantity of field experience itself does not ensure the quality of learning or provide adequate and responsive support that helps teachers improve. From Dewey's [6] perspective, reflection is a process of identifying a problem, considering strategies, and engaging in a recursive process. This foundational perspective has influenced the use of reflective approaches, such as reflection essays or reflective discussions, which are often employed alongside field experiences. These reflective methods also serve as formative assessment tools to monitor and evaluate the complex nature of field experiences and the learning processes of preservice teachers. Feucht et al. [7] highlight that reflection is not merely a theoretical or conceptual endeavor; rather, it includes actions aimed at implementing approaches to solve identified problems. Scholars have emphasized the importance of teacher educators' guidance and support in facilitating pedagogical knowledge and reflections in teacher education [1], [8], [9], [10]. For example, as Shandomo [11] illustrates, reflective practice should be designed for preservice teachers to critically reflect on their experiences with an understanding of themselves, others, and broader curricular and social contexts. Ulusoy [12] also argues that teacher education should teach preservice teachers to analyze the complex and complicated relationships involved in teaching practice for reflection. This is important in moving beyond superficial retrospective discussions of self, mentor teachers, and student-related issues in an individualized manner.

In this study, the emotion and tone of preservice teachers' reflections are analyzed using sentiment analysis with Large Language Model(LLM)s: GPT-4, Gemini, and BERT. We compare the results to understand how each tool categorizes and describes individual reflections and multiple reflections as a whole. This study aims to explore ways to bridge the gaps between qualitative, quantitative, and computational analyses of reflective practices in teacher education. Through the use of sentiment analysis with LLMs, we explore possibilities for nuanced and systemic analysis of preservice teachers' experiences as reflected in their writings. Given the ongoing discussion on the ethics of Artificial Intelligence (AI) [13], [14], [15] surrounding AI development and utilization, we also focus on identifying biases and systemic oppressions that might be inherent in current AI models.

## II. Related Literature Review

Preservice teachers' reflective writing has been analyzed using various methodologies, not limited to certain ontological and epistemological perspectives, ranging from quantitative methods to phenomenological approaches. Zhang et al. [16] consider two broad methodological focuses in the analysis of reflective writing. One approach is manual analysis utilizing a theoretical framework, and the other is computational analysis utilizing technologies. The authors believe that analysis based on a theoretical framework provides a deeper understanding of reflective practice, while computational analysis tends to rely on dictionary definitions and following set rules.

Qualitative analysis of reflections has ranged from discourse analysis [17], case study [18], qualitative interpretive study [19], to phenomenological inquiry [20]. For example, Pedro [19] utilized a qualitative interpretive study to understand the meaning of reflective practice. This study illustrates the learning practice within contexts, interpreting the meaning of the development of professional identities through reflective practice. On the other hand, Yee et al.'s [18] case study, using the six-stage framework, employed several quantitative methods, including sampling and quantifying the results of content analysis.

Research that used quantitative and/or computational analysis includes quantitative content analysis [21], text-mining techniques [22], descriptive statistical analysis [23], and a pre-trained language model (BERT) [24]. For instance, Chen et al. [22] utilized text-mining techniques to analyze broad themes and patterns in preservice teachers' reflections, and the study identified the preservice teachers' interests and focus, as well as the level of reflection based on the lengths of reflections. Poldner et al.'s [21] quantitative content analysis procedure (QCA) study focused on formatively evaluating student teachers' reflective process and compared the thoughts and engagements during two semesters. Zhang et al.'s [16] mixed-method approach, using qualitative and computational analysis, provides insights into a nuanced and multidimensional understanding of preservice teachers, thereby overcoming the shortcomings of existing studies on preservice teacher reflection that rely solely on qualitative or quantitative inquiry.

### III. METHODS

#### A. Data

The written reflections were collected from an elementary teacher education course in the Midwestern United States. These reflections were completed as part of preservice teachers' coursework over a semester, along with their field experience in a local elementary school.

The collected reflections underwent preprocessing by removing titles, reflection prompts, headers, and footers and were then converted into single paragraphs. Each reflection was appended to a row in the Pandas DataFrame. For data persistence, the Pandas DataFrame was stored in the form of a binary pickle file, Python object serialization[25].

#### B. Large language model (LLM)

The LLMs used in this study are GPT-4, Gemini, and BERT.

*1) Generative Pre-trained Transformer-4 (GPT-4):* The GPT-4 is OpenAI's fourth large-scale, multimodal language model and was released on March 14, 2023. For more details, see [26]. This research used the latest GPT-4 version, the gpt-4-turbo-preview model, for sentiment analysis.

*2) Gemini 1.5:* On February 15, 2024, Google DeepMind announced Gemini 1.5, which is a multimodal language model including a sparse mixture-of-experts architecture. For more details, see [27]. This research used the Gemini-1.0-pro model for sentiment analysis.

*3) Bidirectional Encoder Representations from Transformers (BERT):* The research paper for BERT was published by a research team of Google AI Language in October 2018. For more details, see [28]. This research used the TweetNLP model, which is an improved model for sentiment analysis based on BERT [29].

#### C. Sentiment analysis process

Each LLM was called through the provided API. Tone analysis and emotion analysis were performed on the stored pickle file. The prompt instructed by GPT-4 and Gemini to perform sentiment analysis is as follows.

```
/**************************************************
*   instruction = """
*   Your task is to evaluate the level of tone(emotion) of the author in a given reflective journal text.
*   Tone is defined as "tone is the author's attitude toward a subject."
*   (Emotion is defined as "mood is how we are made to feel as readers or the emotion evoked by the author.")
*
*   You should give the text a numeric grade between 0 and 2 for tone(emotion).
*   Tone:
*   2. The text has a positive tone, such as optimism and advertising.
*   1. The text has a neutral tone, such as factual and balanced.
*   0. The text has a negative tone, such as pessimistic and criticizing.
*   Emotion:
*   2. The text has positive emotions, such as joy, anticipation, and happiness.
*   1. The text has neutral emotions, such as surprise, confusion, or nothing.
*   0. The text has negative emotions, such as frustration, fear, anger, disappointment, or outrage.
*
*   [Answer with a two-digit decimal number in the 0-2 range, followed by a semi-colon,
*   and then a brief motivation and keywords with the reason why the author's tone is scored,
*   and then briefly summarize the text, focusing on the tone of the text.
*   For instance: "1.23; The text shows positive tone(emotion) toward a subject matter.
*   These keywords reveal or are linked to the tone (keyword1, keyword2, keyword3, keyword4).
*   And brief summarization." Do not use quotation marks.]
*   """
**************************************************/
```

BERT conducted a reflection analysis by calling the function for tone and motion analysis in the API.

To ensure the independence of sentiment analysis for each reflection, each analysis was performed at a time interval of 2 seconds. In addition, each reflection was randomly selected by the "pandas.DataFrame.sample" function and analyzed to prevent the analysis of the previous reflection from affecting the analysis of the next reflection, as the analysis is performed sequentially. After the analysis for each reflection was completed, all individual reflections were integrated into one text, and the analysis for the entire reflection was performed. In the integration process, to minimize the influence of sequential analysis, the analysis was performed by integration in the order selected in the individual analysis. Tone and emotion analysis using each LLM was repeated five times, and the results were compared.

TABLE I. THE RESULT OF TONE AND EMOTION ANALYSIS OF LLMs FOR OVERALL REFLECTION

| Overall reflection | Model | Score | Motivations |
|---|---|---|---|
| Tone | GPT-4 | 1.5 | The text exhibits a mix of neutral and positive tones |
| Tone | Gemini | 1.89 | The text shows a positive tone toward student potential; however, the author does mention certain drawbacks in the system |
| Tone | BERT | Neutral | Probability of tones 'Negative': 2.4%   'neutral': 88.9%   'positive': 8.7% |
| Emotion | GPT-4 | 0.75 | The text reflects a mix of neutral and negative emotions, with a leaning toward concern and critical reflection rather than outright negativity |
| Emotion | Gemini | 1.5 | The text carries some negative emotions due to frustration, but the author can identify positive aspects and learning experiences |
| Emotion | BERT | Optimism | Probability of emotions 'disgust': 0.6%  'anger': 0.4%  'fear': 0.2%  'optimism': 39.4% 'joy': 17.3%  'love': 0.3%  'sadness': 1.3%  'pessimism': 0.7% 'surprise': 0.9%  'trust': 2.1%  'anticipation': 36.8% |

## IV. RESULTS

### A. Sentiment analysis for overall reflection

Table 1 presents the results for emotion and tone from the sentiment analysis performed on the overall reflections. The assigned scores are 0 for negative, 1 for neutral, and 2 for positive. 'Motivation' represents the result of the written analysis of emotion and tone. For BERT, the output structure differs from that of GPT-4 and Gemini. The model provides the probabilities for each category of emotion and tone.

In tone analysis, GPT-4 represented the data as a mixture of neutral and positive tones with a score of 1.5. Gemini analyzed it as generally positive with a score of 1.89, while BERT identified it as neutral. GPT-4 interpreted the neutral tone as 'the authors reflect on their experiences and observations in the school field and analyze all the challenges and strengths observed in educational practice, student behavior, and the school's community.' The analysis revealed mixed positive tones, indicating 'the authors discuss the possibility of improving the educational approach, understand the positive aspects of the learning environment, and express their anticipation for future class practice.' Gemini highlighted negative aspects of the learning environment and expressed anticipation for future practice, stating 'it refers to the system's shortcomings, such as the emphasis on standardized tests and the lack of time for meaningful math classes.' Meanwhile, the model placed more emphasis on a positive tone, describing 'the text shows a positive tone about the students' potential, and overall, the authors' tone is one of anticipation and optimism.' BERT described the reflections as neutral with a high probability, while both positive and negative tones had low probabilities. Despite subtle differences, we observed that the probability of a positive tone is slightly higher than that of a negative tone.

Emotion analysis revealed significant differences in scores among the three models. GPT-4 assigned a score of 0.75, analyzing a mixture of neutral and negative emotions, whereas Gemini gave a score close to positive at 1.5 points. Gemini emphasized the positive aspects while identifying the presence of some negative emotions. BERT identified optimism as the most prominent emotion.

GPT-4 found that the overall tone combined thoughtful criticism with hopeful insight, reflecting a desire for improvement and adaptation of teaching methods to better meet students' needs and leverage their strengths. However, the model emphasized neutral and negative emotions due to the concerns and critical reflections expressed in the text. Gemini highlighted positive emotions in overall reflections, such as excitement and expectation of teaching and learning. The model also mentioned negative emotions, specifically frustration and fear related to high-stakes testing and students' struggles with it. BERT captured feelings of optimism and anticipation, indicating them as having the highest probability.

### B. Sentiment analysis for individual reflection

TABLE II. THE RESULT OF TONE AND EMOTION ANALYSIS OF LLMs FOR AN INDIVIDUAL REFLECTION OF EACH TRIAL

| Emotion/Tone | Trial 1 | Trial 2 | Trial 3 | Trial 4 | Trial 5 |
|---|---|---|---|---|---|
| GPT-4 | 0.75/1.5 | 1.5/1.67 | 1.5/1.5 | 1.0/1.5 | 1.5/1.5 |
| Gemini | 1.3/1.0 | 2.7/1.33 | 1.5/1.5 | 1.3/1.0 | 1.5/1.0 |
| BERT | Optimism 84.2% / Positive 54.9% | Optimism 84.2% / Positive 54.9% | Optimism 84.2% / Positive 54.9% | Optimism 84.2% / Positive 54.9% | Optimism 84.2% / Positive 54.9% |

We also conducted a sentimental analysis of each individual reflection. Table 2 presents the results of each trial of the sentiment analysis performed on one of the individual reflections. Overall, all models analyzed this reflection as having a positive tone and emotion. Through this analysis, we confirmed the consistency of the model results across trials. Although GPT-4 initially indicated negative emotion in the first trial, it consistently showed a positive tone and emotion in subsequent trials. We observed significant variations in Gemini's scores across trials compared to other models. In tone analysis, Gemini's scores were predominantly neutral or positive. In emotion analysis, the results were generally positive, with scores ranging from 1.3 to 2.7. In addition, we consider the score of 2.7 to be abnormal as it falls outside the range indicated by the prompt. BERT's results were relatively consistent across trials compared to other models.

## V. Discussion

In comparison, GPT-4 and Gemini assigned similar scores for tone but different scores for emotion. However, despite numerical discrepancies, descriptive analysis results indicated that both models produced similar outcomes in tone and emotion. In other words, while the two models may emphasize different texts, their overall analyses are generally similar. We view that BERT is more suitable for expressing the overall tone and emotion of text in understandable numbers and probabilities, but it is weaker in descriptive analysis due to limited result description. However, BERT's emotion analysis result, indicating 'the probability of optimism and anticipation was the highest,' aligns with Gemini's tone analysis result, stating 'the authors' tone is one of anticipation and optimism.' This suggests that the results from all three models are generally consistent.

### A. Scores of tone and emotion

In the analysis conducted for each model, performed five times, the score expressed on a scale showed a somewhat different form of expression compared to human common sense. In the prompt provided for the analysis, a negative tone or emotion was to be rated on a scale of 0 for negative, 1 for neutral, and 2 for positive. Humans tend to understand tone and emotion as a spectrum, meaning they would likely assign a rating close to 2 for positive texts and close to 0 for negative ones. However, we found that LLM's scoring approach differs from human perception and understanding of emotion and tone.

The results indicate that LLM analyzes the similarity or probability of each positive, neutral, and negative by treating them as distinct categories rather than as a spectrum. LLMs determine that each positive, neutral, and negative emotion is an independent category and do not recognize a concept of neutrality between the negative and positive categories. This finding suggests that BERT's output form, which provides probabilities for each emotion, is a more suitable output form for sentiment analysis using LLM.

### B. Fine-tuned model for sentiment analysis for reflections of preservice teachers

This study employed a pre-trained model because the quantity of reflection data for the sentiment analysis was insufficient to fine-tune LLMs. To successfully integrate LLM analysis into teacher education, it is crucial to develop an analysis approach and result format that are comprehensive and relevant for preservice teachers and teacher educators. As indicated by Gemini's results, LLMs need to be fine-tuned for specific contexts and purposes, aligning with the perspectives and directions pursued in teacher education. Considering that the results from BERT, which specializes in sentiment analysis based on data from Twitter, were relatively consistent, we can anticipate that Gemini's results could become more consistent and relevant as it is fine-tuned with a large dataset of reflections from preservice teachers.

## VI. Conclusion

Given the key role of reflection in teacher education, it is important for teacher educators to provide timely and adequate feedback on the reflective process. Our study has shown that LLMs have the potential to support preservice teachers in developing their professional practice within the context of larger social discourse and the reflective processes of others. However, we recognize the pitfall of LLMs in that they simplify and miss nuances when it comes to the complexity of the lived experiences of preservice teachers. For example, if an analysis focuses solely on literal meaning and numerical domination of certain emotions and sentiments labeled as negative, concerning standardized testing and the classroom/school environment, it will miss the nuance of motivation, inspiration, and hope for their future practice. In contrast, the domination of positive emotions and sentiments does not necessarily mean that preservice teachers are developing pedagogical practices responsive to their students and social contexts. When one is analyzing the process of becoming an educator, it is crucial for an analytical model and tool to avoid reaching hasty conclusions while still being able to capture the nuances and relational aspects of teaching practice and the larger social and educational contexts.